\documentclass[letterpaper]{article} % DO NOT CHANGE THIS
\usepackage{aaai25}  % DO NOT CHANGE THIS
\usepackage{times}  % DO NOT CHANGE THIS
\usepackage{helvet}  % DO NOT CHANGE THIS
\usepackage{courier}  % DO NOT CHANGE THIS
\usepackage[hyphens]{url}  % DO NOT CHANGE THIS
\usepackage{graphicx} % DO NOT CHANGE THIS
\usepackage{amsmath}
\usepackage{amsthm}
\usepackage{booktabs}
\usepackage{tabularx}
\usepackage{algorithm,algpseudocode}
\usepackage[switch]{lineno}
\usepackage{subfig}
\usepackage{mathtools}
\usepackage{xcolor}

\usepackage{tabularx}  
\newcolumntype{L}[1]{>{\raggedright\arraybackslash}m{#1}}
\newcolumntype{P}[1]{>{\centering\arraybackslash}p{#1}}
\usepackage{amssymb}
\usepackage{natbib}  % DO NOT CHANGE THIS AND DO NOT ADD ANY OPTIONS TO IT
\usepackage{caption} % DO NOT CHANGE THIS AND DO NOT ADD ANY OPTIONS TO IT
\usepackage{newfloat}
\usepackage{listings}
\DeclareCaptionStyle{ruled}{labelfont=normalfont,labelsep=colon,strut=off} % DO NOT CHANGE THIS
\floatstyle{ruled}
\newfloat{listing}{tb}{lst}{}
\floatname{listing}{Listing}
\nocopyright %-- Your paper will not be published if you use this command
\title{Anytime Multi-Agent Path Finding with an Adaptive Delay-Based Heuristic}
\author{
    Thomy Phan\equalcontrib\textsuperscript{\rm 1}, Benran Zhang\equalcontrib\textsuperscript{\rm 1}, Shao-Hung Chan\textsuperscript{\rm 1}, Sven Koenig\textsuperscript{\rm 2}
}
\affiliations{
    \textsuperscript{\rm 1}University of Southern California\\
    \textsuperscript{\rm 2}University of California, Irvine\\
	\{thomy.phan,jimyzhan,shaohung\}@usc.edu, sven.koenig@uci.edu
}

\begin{document}

\maketitle

\begin{abstract}
 Anytime \emph{multi-agent path finding (MAPF)} is a promising approach to scalable and collision-free path optimization in multi-agent systems. MAPF-LNS, based on \emph{Large Neighborhood Search (LNS)}, is the current state-of-the-art approach where a fast initial solution is iteratively optimized by destroying and repairing selected paths of the solution. Current MAPF-LNS variants commonly use an adaptive selection mechanism to choose among multiple destroy heuristics. However, to determine promising destroy heuristics, MAPF-LNS requires a considerable amount of exploration time. As common destroy heuristics are stationary, i.e., non-adaptive, any performance bottleneck caused by them cannot be overcome by adaptive heuristic selection alone, thus limiting the overall effectiveness of MAPF-LNS.
In this paper, we propose \emph{Adaptive Delay-based Destroy-and-Repair Enhanced with Success-based Self-learning (ADDRESS)} as a single-destroy-heuristic variant of MAPF-LNS. ADDRESS applies restricted Thompson Sampling to the top-$K$ set of the most delayed agents to select a seed agent for adaptive LNS neighborhood generation. We evaluate ADDRESS in multiple maps from the MAPF benchmark set and demonstrate cost improvements by at least 50\% in large-scale scenarios with up to a thousand agents, compared with the original MAPF-LNS and other state-of-the-art methods.
\end{abstract}

\section{Introduction}

A wide range of real-world applications like goods transportation in warehouses,  search and rescue missions,  and traffic management can be formulated as \emph{Multi-Agent Path Finding (MAPF)} problem,  where the goal is to find collision-free paths for multiple agents with each having an assigned start and goal location. Finding optimal solutions, w.r.t. minimal flowtime or makespan is NP-hard, which limits scalability of most state-of-the-art MAPF solvers \cite{ratner1986finding,sharon2012conflict,yu2013structure}.

\emph{Anytime MAPF} based on \emph{Large Neighborhood Search (LNS)} is a promising approach to finding fast and high-quality solutions to the MAPF problem within a fixed time budget \cite{li2021anytime}. Given an initial feasible solution and a set of destroy heuristics, LNS iteratively destroys and replans a fixed number of paths, according to an agent \emph{neighborhood}, until the time budget runs out. MAPF-LNS represents the current state-of-the-art in anytime MAPF and has been experimentally shown to scale up to scenarios with hundreds of agents \cite{li2021anytime}. Due to its increasing popularity, several extensions have been proposed like fast local repairing, integration of primal heuristics, machine learning-guided neighborhood selection, neighborhood size adaptation, and parallelism \cite{chan2024anytime,HuangAAAI22,LamICAPS23,li2022lns2,phanAAAI24}.

\begin{figure}
	\centering
	\includegraphics[width=0.3\textwidth]{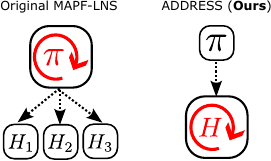}
     \caption{Scheme of our contribution. Instead of using an adaptive selection mechanism $\pi$ to choose among multiple stationary destroy heuristics $H_x$ \cite{li2021anytime}, ADDRESS (our approach) only uses a \emph{single adaptive heuristic}.}
     \label{fig:address_high_level}
\end{figure}

Current MAPF-LNS variants use an \emph{adaptive selection mechanism} $\pi$ to choose from the set of destroy heuristics, as illustrated in Figure \ref{fig:address_high_level} \cite{ropke2006adaptive}. However, to determine promising destroy heuristics, MAPF-LNS requires a considerable amount of exploration time. As common destroy heuristics are \emph{stationary}, i.e., non-adaptive \cite{li2021anytime}, any performance bottleneck caused by them cannot be overcome by the adaptive selection mechanism $\pi$ alone, thus limiting the overall effectiveness of MAPF-LNS.

In this paper, we propose \emph{Adaptive Delay-based Destroy-and-Repair Enhanced with Success-based Self-learning (ADDRESS)}, as a single-destroy-heuristic variant of MAPF-LNS, illustrated in Figure \ref{fig:address_high_level}. ADDRESS applies restricted Thompson Sampling to the top-$K$ set of the most delayed agents to select a seed agent for adaptive LNS neighborhood generation. Our contributions are as follows:
\begin{itemize}
\item We discuss a performance bottleneck of the current empirically most effective destroy heuristic in MAPF-LNS and its implications for large-scale scenarios.
\item We define an adaptive destroy heuristic, called ADDRESS heuristic, to generate neighborhoods based on the top-$K$ set of the most delayed agents, using multi-armed bandits like Thompson Sampling. We formulate a simplified variant of MAPF-LNS using only our ADDRESS heuristic, as illustrated in Figure \ref{fig:address_high_level}.
\item We evaluate ADDRESS in multiple maps from the MAPF benchmark set \cite{stern2019multi} and demonstrate cost improvements by at least 50\% in large-scale scenarios with up to a thousand agents, compared with the original MAPF-LNS and other state-of-the-art methods.% When combined with the original MAPF-LNS, our ADDRESS heuristic is consistently preferred over all other destroy heuristics.
\end{itemize}

While our paper focuses on MAPF, our ADDRESS heuristic can also be applied to other problem classes, where variables can be sorted by their cost contribution to generate LNS neighborhoods \cite{pisinger2019large}.

\section{Background}

\subsection{Multi-Agent Path Finding (MAPF)}

We focus on \emph{maps} as undirected unweighted \emph{graphs} $G = \langle \mathcal{V}, \mathcal{E} \rangle$, where vertex set $\mathcal{V}$ contains all possible locations and edge set $\mathcal{E}$ contains all possible transitions or movements between adjacent locations. An \emph{instance} $I$ consists of a map $G$ and a set of \emph{agents} $\mathcal{A} = \{a_1, ..., a_m\}$ with each agent $a_i$ having a \emph{start location} $s_i \in \mathcal{V}$ and a \emph{goal location} $g_i \in \mathcal{V}$. At every time step $t$, all agents can move along the edges in $\mathcal{E}$ or wait at their current location \cite{stern2019multi}.

MAPF aims to find a collision-free plan for all agents. A \emph{plan} $P = \{ p_1, ..., p_m \}$ consists of individual paths $p_i = \langle p_{i,1}, ..., p_{i,l(p_i)} \rangle$ per agent $a_i$, where $\langle p_{i,t}, p_{i,t+1} \rangle = \langle p_{i,t+1}, p_{i,t} \rangle \in \mathcal{E}$, $p_{i,1} = s_i$, $p_{i,l(p_i)} = g_i$, and $l(p_i)$ is the \emph{length} or \emph{travel distance} of path $p_i$. The \emph{delay} $\textit{del}(p_i)$ of path $p_i$ is defined by the difference of path length $l(p_i)$ and the length of the shortest path from $s_i$ to $g_i$ w.r.t. map $G$.

In this paper, we consider \emph{vertex conflicts} $\langle a_i, a_j, v, t \rangle$ that occur when two agents $a_i$ and $a_j$ occupy the same location $v \in \mathcal{V}$ at time step $t$ and \emph{edge conflicts} $\langle a_i, a_j, u, v, t \rangle$ that occur when two agents $a_i$ and $a_j$ traverse the same edge $\langle u, v \rangle \in \mathcal{E}$ in opposite directions at time step $t$ \cite{stern2019multi}. A plan $P$ is a \emph{solution}, i.e., \emph{feasible} when it does not have any vertex or edge conflicts. Our goal is to find a feasible solution by minimizing the \emph{flowtime} $\sum_{p \in P} l(p)$ equivalent to minimizing the \emph{sum of delays} or \emph{(total) cost} $c(P) = \sum_{p \in P} \textit{del}(p)$. In the context of anytime MAPF, we also consider the \emph{Area Under the Curve (AUC)} as a measure of how quickly we approach the quality of our final solution.

\subsection{Anytime MAPF with LNS}\label{subsec:mapf_lns_background}

\emph{Anytime MAPF} searches for solutions within a given \emph{time budget}. The solution quality monotonically improves with increasing time budget \cite{cohen2018anytime,li2021anytime}.

\emph{MAPF-LNS} based on \emph{Large Neighborhood Search (LNS)} is the current state-of-the-art approach to anytime MAPF and shown to scale up to large-scale scenarios with hundreds of agents \cite{HuangAAAI22,li2021anytime}. Starting with an initial feasible plan $P$, e.g., found via \emph{prioritized planning (PP)} from \cite{silver2005cooperative}, MAPF-LNS iteratively modifies $P$ by destroying $N < m$ paths of the \emph{neighborhood} $A_N \subset \mathcal{A}$. The destroyed paths $P^{-} \subset P$ are then repaired or replanned using PP to quickly generate new paths $P^{+}$. If the new cost $c(P^{+})$ is lower than the previous cost $c(P^{-})$, then $P$ is replaced by $(P \backslash P^{-}) \cup P^{+}$, and the search continues until the time budget runs out. The result of MAPF-LNS is the last accepted solution $P$, with the lowest cost so far.

MAPF-LNS uses a set of three \emph{destroy heuristics}, namely a \emph{random uniform selection} of $N$ agents, an \emph{agent-based heuristic}, and a \emph{map-based heuristic} \cite{li2021anytime}. The agent-based heuristic generates a neighborhood, including a \emph{seed agent} $a_j$ with the current maximum delay and other agents, determined via random walks, that prevent $a_j$ from achieving a lower delay. The map-based heuristic randomly chooses a vertex $v \in \mathcal{V}$ with a degree greater than 2 and generates a neighborhood of agents moving around $v$. All heuristics are randomized but \emph{stationary} since they do not adapt their rules and degree of randomization, i.e., the distributions, based on prior improvements made to the solution.

The original MAPF-LNS uses an adaptive \emph{selection mechanism} $\pi$ by maintaining selection weights to choose destroy heuristics $P$ \cite{li2021anytime,ropke2006adaptive}.

\subsection{Multi-Armed Bandits}
\emph{Multi-armed bandits (MABs)} or simply bandits are fundamental decision-making problems, where an \emph{MAB or selection algorithm} $\pi$ repeatedly chooses an \emph{arm} $j$ among a given set of arms or \emph{options} $\{1, ..., K\}$ to maximize an expected \emph{reward} of a stochastic reward function $\mathcal{R}(j) := X_{j}$, where $X_{j}$ is a random variable with an unknown distribution $f_{X_{j}}$ \cite{auer2002finite}. To solve an MAB, one has to determine an \emph{optimal arm} $j^{*}$, which maximizes the expected reward $\mathbb{E}\big[X_{j}\big]$. The MAB algorithm $\pi$ has to balance between exploring all arms $j$ to accurately estimate $\mathbb{E}\big[X_{j}\big]$ and exploiting its knowledge by greedily selecting the arm $j$ with the currently highest estimate of $\mathbb{E}\big[X_{j}\big]$. This is known as the \emph{exploration-exploitation dilemma}, where exploration can find arms with higher rewards but requires more time for trying them out, while exploitation can lead to fast convergence but possibly gets stuck in a poor local optimum. We will focus on \emph{Thompson Sampling} and $\epsilon$\emph{-Greedy} as MAB algorithms and explain them in Section \ref{subsec:address_heuristic}.

\section{Related Work}

\subsection{Multi-Armed Bandits for LNS}
In recent years, MABs have been used to tune learning and optimization algorithms on the fly \cite{badia2020agent57,hendel2022adaptive,schaul2019adapting}. UCB1 and $\epsilon$-Greedy are commonly used for LNS destroy heuristic selection in \emph{traveling salesman problems (TSP)}, \emph{mixed integer linear programming (MILP)}, and \emph{vehicle routing problems (VRP)} \cite{Chen2016AMB,hendel2022adaptive}. In most cases, a heavily engineered reward function with several weighted terms is used for training the MAB. Recently, a MAPF-LNS variant, called BALANCE, has been proposed to adapt the neighborhood size $N$ along with the destroy heuristic choice using a bi-level Thompson Sampling approach \cite{phanAAAI24}.

Instead of adapting the destroy heuristic selection, we propose a \emph{single adaptive destroy heuristic}, thus \emph{simplifying} the high-level MAPF-LNS procedure (Figure \ref{fig:address_high_level}). Our destroy heuristic uses \emph{restricted Thompson Sampling} with \emph{simple binary rewards} to select a seed agent from the \emph{top-}$K$ \emph{set of the most delayed agents} for LNS neighborhood generation, which can also be applied to other problem classes, such as TSP, MILP, or VRP \cite{pisinger2019large}.

\subsection{Machine Learning in Anytime MAPF}

Machine learning has been used in MAPF to directly learn collision-free path finding, to guide the node selection in search trees, or to select appropriate MAPF algorithms for certain maps \cite{alkazzi2024comprehensive,huang2021learning,kaduri2020algorithm,phan2024confidence,phan2025confidence,sartoretti2019primal}. \cite{HuangAAAI22,yan2024neural} propose machine learning-guided variants of MAPF-LNS, where neighborhoods are generated by stationary procedures, e.g., the destroy heuristics of \cite{li2021anytime}. The neighborhoods are then selected via an offline trained model. Such methods cannot adapt during the search and require extensive prior efforts like data acquisition, model training, and feature engineering.

We focus on \emph{adaptive} approaches to MAPF-LNS using \emph{online learning via MABs}. Our destroy heuristic can adjust on the fly via \emph{binary reward signals}, indicating a successful or failed improvement of the solution quality. The rewards are \emph{directly} obtained from the LNS without any prior data acquisition or expensive feature engineering.

\section{Adaptive Delay-Based MAPF-LNS}

We now introduce \emph{Adaptive Delay-based Destroy-and-Repair Enhanced with Success-based Self-learning (ADDRESS)} as a simplified yet effective variant of MAPF-LNS.

\subsection{Original Agent-Based Destroy Heuristic}\label{subsec:original_heuristic}

Our adaptive destroy heuristic is inspired by the agent-based heuristic of \cite{li2021anytime}, which is empirically confirmed to be the most effective standalone heuristic in most maps \cite{li2021anytime,phanAAAI24}.

The idea is to select a \emph{seed agent} $a_j \in \mathcal{A}$, whose path $p_j \in P$ has a high potential to be shortened, indicated by its delay $\textit{del}(p_j)$. A random walk is performed from a random position in $p_j$ to collect $N-1$ other agents $a_i$ whose paths $p_i$ are crossed by the random walk, indicating their contribution to the delay $\textit{del}(p_j)$, to generate a neighborhood $A_N \subset \mathcal{A}$ of size $|A_N| = N < m$ for LNS destroy-and-repair.

The original destroy heuristic of \cite{li2021anytime} greedily selects the seed agent with the maximum delay $\textit{max}_{p_i \in P}\textit{del}(p_i)$. To avoid repeated selection of the same agent, the original heuristic maintains a \emph{tabu list}, which is emptied when all agents have been selected or when the current seed agent $a_j$ has no delay, i.e., $\textit{del}(p_j) = 0$. Therefore, the heuristic has to iterate over all agents $a_i \in \mathcal{A}$ in the worst case, which is time-consuming for large-scale scenarios with many agents, introducing a potential performance bottleneck. The original MAPF-LNS cannot overcome this bottleneck because it only adapts the high-level heuristic selection via $\pi$, as shown in Figure \ref{fig:address_high_level}, and thus can only switch to other (less effective) destroy heuristics as an alternative.

\subsection{ADDRESS Destroy Heuristic}\label{subsec:address_heuristic}

Our goal is to overcome the limitation of the original agent-based destroy heuristic, and consequently of MAPF-LNS, using MABs. We model each agent $a_i \in \mathcal{A}$ as an arm $i$ and maintain two counters per agent, namely $\alpha_i > 0$ for \emph{successful cost improvements}, and $\beta_i > 0$ for \emph{failed cost improvements}. Both counters represent the parameters of a Beta distribution $\textit{Beta}(\alpha_i,\beta_i)$, which estimates the potential of an agent $a_i \in \mathcal{A}$ to improve the solution as a seed agent. $\textit{Beta}(\alpha_i,\beta_i)$ has a mean of $\frac{\alpha_i}{\alpha_i + \beta_i}$ and is initialized with $\alpha_i = 1$ and $\beta_i = 1$, corresponding to an initial 50:50 chance estimate that an agent $a_i$ could improve the solution if selected as a seed agent \cite{chapelle2011empirical}.

Since the number of agents $m$ can be large, a naive MAB would need to explore an enormous arm space, which poses a similar bottleneck as the tabu list approach of the original agent-based heuristic (Section \ref{subsec:original_heuristic}).
Thus, we restrict the agent selection to the \emph{top-}$K$ \emph{set} $\mathcal{A}_K \subseteq \mathcal{A}$ of the most delayed agents with $K \leq m$ to ease exploration.

The simplest MAB is $\epsilon$\emph{-Greedy}, which selects a random seed agent $a_i \in \mathcal{A}_K$ with a probability of $\epsilon \in [0,1]$, and the agent with the highest expected success rate $\frac{\alpha_i}{\alpha_i + \beta_i}$ with the complementary probability of $(1-\epsilon)$.

We propose a \emph{restricted Thompson Sampling} approach to select a seed agent from $\mathcal{A}_K$. For each agent $a_i \in \mathcal{A}_K$ within the top-$K$ set, we sample an estimate $q_i \sim \textit{Beta}(\alpha_i,\beta_i)$ of the solution improvement rate and select the agent with the highest sampled estimate $q_i$. Thompson Sampling is a Bayesian approach with $\textit{Beta}(1, 1)$ being the \emph{prior distribution} of the improvement success rate and $\textit{Beta}(\alpha_i,\beta_i)$ with updated parameters $\alpha_i$ and $\beta_i$ being the \emph{posterior distribution} \cite{chapelle2011empirical,thompson1933likelihood}.

Our destroy heuristic, called ADDRESS heuristic, first sorts all agents w.r.t. their delays to determine the top-$K$ set $\mathcal{A}_K \subseteq \mathcal{A}$ of the most delayed agents. Restricted Thompson Sampling is then applied to the parameters $\alpha_i$ and $\beta_i$ of all agents $a_i \in \mathcal{A}_K$ to select a seed agent $a_j$. An LNS neighborhood $A_N \subset \mathcal{A}$ is generated via random walks, according to \cite{li2021anytime}, by adding agents $a_i \in \mathcal{A}$ whose paths are crossed by the random walk. Note that these agents $a_i \in A_N \backslash \{a_j\}$ are not necessarily part of the top-$K$ set $\mathcal{A}_K$.

The full formulation of our ADDRESS heuristic with Thompson Sampling is provided in Algorithm \ref{algorithm:ADDRESSHeuristic}, where $I$ represents the instance to be solved, $P$ represents the current solution, $K$ restricts the seed agent selection, and $\langle \alpha_i, \beta_i \rangle_{1 \leq i \leq m}$ represent the parameters for the corresponding Beta distributions per agent for Thompson Sampling.

\begin{algorithm}
\caption{ADDRESS Destroy Heuristic}\label{algorithm:ADDRESSHeuristic}
\begin{algorithmic}[1]
\Procedure{$\textit{ADDRESSDestroy}(I, P, K, \langle \alpha_i, \beta_i \rangle_{1 \leq i \leq m})$}{}
\State Sort all agents $a_i \in \mathcal{A}$ w.r.t. their delays $\textit{del}(p_i)$
\State Select the top-$K$ set $\mathcal{A}_K \subseteq \mathcal{A}$ w.r.t. the delays
\For{agent $a_i$ in $\mathcal{A}_K$}
\State $q_i \sim \textit{Beta}(\alpha_i,\beta_i)$ \Comment{Restr. Thompson Sampling}
\EndFor
\State $j \leftarrow \textit{argmax}_{i}q_i$ \Comment{Select the seed agent index}
\State $A_N \sim \textit{RandomWalkNeighborhood}(I, P, a_j)$ \Comment{Random walk routine of \cite{li2021anytime}}
\State\Return $\langle A_N, j \rangle$ \Comment{Neighborhood and seed agent}
\EndProcedure
\end{algorithmic}
\end{algorithm}

\subsection{ADDRESS Formulation}

We now integrate our ADDRESS heuristic into the MAPF-LNS algorithm \cite{li2021anytime}. For a more focused search, we propose a simplified variant, called ADDRESS, which only uses our adaptive destroy heuristic instead of determining a promising stationary heuristic via time-consuming exploration, as illustrated in Figure \ref{fig:address_high_level}.

\begin{figure*}
	\centering
	\includegraphics[width=0.8\textwidth]{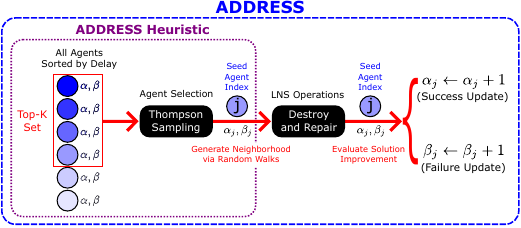}
     \caption{Detailed overview of ADDRESS. For each agent $a_i \in \mathcal{A}$, we maintain two parameters $\alpha_i, \beta_i > 0$. At each LNS iteration, all agents are sorted w.r.t. to their delays. A restricted Thompson Sampling approach is applied to the top-$K$ set of the most delayed agents, according to their samples $q_i \sim \textit{Beta}(\alpha_i,\beta_i)$, to choose a \emph{seed agent index} $j$. The path of the seed agent $a_j$ is used to generate an LNS neighborhood $A_N \subset \mathcal{A}$ via random walks. After running the LNS destroy-and-repair operations on $A_N$, the parameters $\alpha_j$ or $\beta_j$ of the seed agent $a_j$ are updated, depending on the cost improvement of the new solution.}
     \label{fig:address_scheme}
\end{figure*}

ADDRESS iteratively invokes our proposed destroy heuristic of Algorithm \ref{algorithm:ADDRESSHeuristic} with the parameters $\langle \alpha_i, \beta_i \rangle_{1 \leq i \leq m}$ to select a seed agent $a_j \in \mathcal{A}$ and generate an LNS neighborhood $A_N \subset \mathcal{A}$ using the random walk procedure of the original MAPF-LNS \cite{li2021anytime}. Afterward, the standard destroy-and-repair operations of MAPF-LNS are performed on the neighborhood $A_N$ to produce a new solution $P' = (P \backslash P^{-}) \cup P^{+}$. If the new solution $P'$ has a lower cost than the previous solution $P$, then $\alpha_j$ is incremented and $P$ is replaced by $P'$. Otherwise, $\beta_j$ is incremented. The whole procedure is illustrated in Figure \ref{fig:address_scheme}.

The full formulation of ADDRESS is provided in Algorithm \ref{algorithm:ADDRESS}, where $I$ represents the instance to be solved and $K$ restricts the seed agent selection. The parameters $\langle \alpha_i, \beta_i \rangle_{1 \leq i \leq m}$ are all initialized with 1 as a uniform prior.

\subsection{Conceptual Discussion}\label{subsec:conceptual_discussion}

ADDRESS is a simple and adaptive approach to scalable anytime MAPF. The adaptation is controlled by the learnable parameters $\alpha_i$ and $\beta_i$ per agent $a_i$, and the top-$K$ ranking of potential seed agents. Our ADDRESS heuristic can significantly improve MAPF-LNS, overcoming the performance bottleneck of the original agent-based heuristic of \cite{li2021anytime} by selecting seed agents via MABs instead of greedily, and restricting the selection to the top-$K$ set of the most delayed agents $\mathcal{A}_K$ to ease exploration.

The parameters $\alpha_i$ and $\beta_i$ enable the seed agent selection via Thompson Sampling, which considers the improvement success rate under uncertainty via Bayesian inference \cite{thompson1933likelihood}. Unlike prior MAB-enhanced LNS approaches, ADDRESS only uses binary rewards denoting success or failure, thus greatly simplifying our approach compared to alternative MAB approaches \cite{Chen2016AMB,chmiela2023online,hendel2022adaptive,phanAAAI24}.

The top-$K$ set enables efficient learning by reducing the number of options for Thompson Sampling, which otherwise would require exhaustive exploration of all agents $a_i \in \mathcal{A}$. The top-$K$ set supports fast adaptation by filtering out seed agent candidates whose paths were significantly shortened earlier. While the top-$K$ ranking causes some overhead due to sorting agents, our experiments in Section \ref{sec:experiments} suggest that the sorting overhead is outweighed by the performance gains regarding cost and AUC in large-scale scenarios.

Our single-destroy-heuristic approach enables a more focused search toward high-quality solutions without time-consuming exploration of stationary (and less effective) destroy heuristics. Due to its simplicity, our ADDRESS heuristic can be easily applied to other problem classes, such as TSP, MILP, or VRP, when using so-called \emph{worst} or \emph{critical destroy heuristics}, focusing on high-cost variables that ``spoil" the structure of the solution \cite{pisinger2019large}. We defer such applications to future work.

\begin{algorithm}[ht!]
\caption{MAPF-LNS with our ADDRESS Heuristic}\label{algorithm:ADDRESS}
\begin{algorithmic}[1]
\Procedure{$\textit{ADDRESS}(I,K)$}{}
\State $\langle \alpha_i, \beta_i \rangle \leftarrow \langle 1, 1 \rangle$ for all agents $a_i \in \mathcal{A}$
\State $P = \{ p_1, ..., p_m \} \leftarrow \textit{RunInitialSolver(I)}$
\While{runtime limit not exceeded}
\State $B \leftarrow \langle \alpha_i, \beta_i \rangle_{1 \leq i \leq m}$ \Comment{Distribution parameters}
\State $\langle A_N, j \rangle \leftarrow \textit{ADDRESSDestroy}(I, P, K, B)$ \Comment{See Algorithm \ref{algorithm:ADDRESSHeuristic}}
\State $P^{-} \leftarrow \{p_i | a_i \in A_N\}$
\State $P^{+} \leftarrow \textit{DestroyAndRepair}(I, A_N, P \backslash P^{-})$
\If{$c(P^{-}) - c(P^{+}) > 0$}
\State $P \leftarrow (P \backslash P^{-}) \cup P^{+}$ \Comment{Replace solution}
\State $\alpha_j \leftarrow \alpha_j + 1$ \Comment{Success update}
\Else
\State $\beta_j \leftarrow \beta_j + 1$ \Comment{Failure update}
\EndIf
\EndWhile
\State\Return $P$
\EndProcedure
\end{algorithmic}
\end{algorithm}

\section{Experiments\footnote{Code is provided at \url{https://github.com/JimyZ13/ADDRESS}.}}\label{sec:experiments}

\paragraph{Maps}

We evaluate ADDRESS on five maps from the MAPF benchmark set of \cite{stern2019multi}, namely (1) a \texttt{Random} map (\emph{Random-32-32-20}), (2) two \texttt{Game} maps \emph{Ost003d} and (3) \emph{Den520d}, (4) a \texttt{Warehouse} map (\emph{Warehouse-20-40-10-2-2}), and (5) a \texttt{City} map (\emph{Paris\_1\_256}). All maps have different sizes and structures. We conduct all experiments on the publicly available 25 random scenarios per map.

\paragraph{Anytime MAPF Algorithms} We implemented ADDRESS with Thompson Sampling and $\epsilon$-Greedy, denoted by \textit{ADDRESS (X)}, where \emph{X} is the MAB algorithm. Our implementation is based on the public code of \cite{li2022lns2,phanAAAI24}. We use the original MAPF-LNS, MAPF-LNS2, and BALANCE implementations from the respective code bases with their default configurations, unless stated otherwise. We also run LaCAM* from \cite{okumura2023lacam2}.

We always set the neighborhood size $N = 8$ (except for BALANCE, which automatically adapts $N$), $K = 32$, and use Thompson Sampling for ADDRESS and BALANCE, unless stated otherwise. $\epsilon$-Greedy is used with $\epsilon = \frac{1}{2}$. All MAPF-LNS variants use PP to generate initial solutions and repair LNS neighborhoods, as suggested in \cite{li2021anytime,HuangAAAI22}.

\paragraph{Compute Infrastructure}

All experiments were run on a high-performance computing cluster with CentOS Linux, Intel Xeon 2640v4 CPUs, and 64 GB RAM.

\subsection{Experiment -- Choice of $K$}\label{subsec:k_choice}

\paragraph{Setting} We run ADDRESS with Thompson Sampling and $\epsilon$-Greedy to evaluate different choices of $K \in \{8, 16, 32, 64, 128, 256\}$ on the \texttt{Den520d} and \texttt{City} map with $m = 700$ agents and a time budget of 60 seconds. The results are compared with MAPF-LNS using only the agent-based heuristic of \cite{li2021anytime}, as a stationary variant.

\begin{figure}
	\centering
	\includegraphics[width=0.47\textwidth]{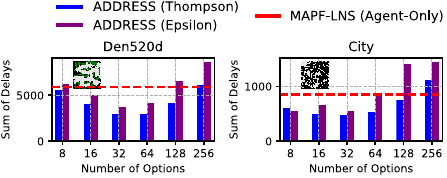}
     \caption{Sum of delays for ADDRESS (using $\epsilon$-greedy or Thompson Sampling) compared with MAPF-LNS (using only the agent-based heuristic) for different numbers of options $K$ with $m = 700$ agents in both maps, a time budget of 60 seconds, and $\epsilon = \frac{1}{2}$.}
     \label{fig:topk_evaluation}
\end{figure}

\paragraph{Results} The results are shown in Figure \ref{fig:topk_evaluation}. ADDRESS with Thompson Sampling always performs best when $K < 256$. However, ADDRESS is more sensitive to $K$ when using $\epsilon$-Greedy, which only outperforms the original agent-based heuristic, when $8 < K < 64$. In all our test maps, both ADDRESS variants work best when $K = 32$.

\paragraph{Discussion} The results indicate that both ADDRESS variants with either Thompson Sampling or $\epsilon$-Greedy can notably outperform the original agent-based heuristic of MAPF-LNS with sufficient restriction via $K < m$. Thompson Sampling is more robust regarding the choice of $K$.

\subsection{Experiment -- Delay-Based Heuristics}

\paragraph{Setting} Next, we evaluate the search progress of ADDRESS with Thompson Sampling and $\epsilon$-Greedy for different time budgets on the \texttt{Den520d} and \texttt{City} map with $m = 700$ agents. The results are compared with MAPF-LNS using only the agent-based heuristic, as a stationary variant.

\begin{figure}
	\centering
	\includegraphics[width=0.47\textwidth]{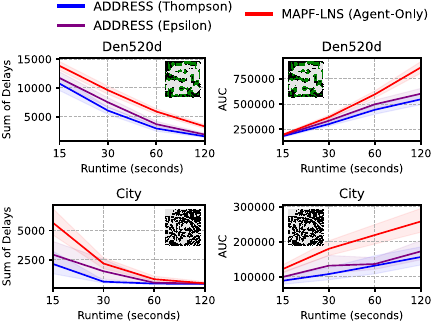}
     \caption{Sum of delays and AUC for ADDRESS (using $\epsilon$-greedy or Thompson Sampling) compared with MAPF-LNS (using only the agent-based heuristic) for different time budgets (\underline{starting from 15 seconds}) with $m = 700$ agents in both maps and $\epsilon = \frac{1}{2}$. Shaded areas show the 95\% confidence interval.}
     \label{fig:single_heuristic_runtime}
\end{figure}

\paragraph{Results} The results are shown in Figure \ref{fig:single_heuristic_runtime}. Both ADDRESS variants outperform the agent-based MAPF-LNS by always achieving lower sums of delays and AUC values, which indicate that ADDRESS always improves faster than the original agent-based heuristic. Thompson Sampling always performs at least as well as $\epsilon$-Greedy.

\paragraph{Discussion} The results demonstrate the potential of both ADDRESS variants to improve MAPF-LNS over the original agent-based heuristic for any time budget w.r.t. solution cost and speed of cost improvement. This confirms that the combination of MABs and the top-$K$ set can overcome the performance bottleneck of the original agent-based heuristic (Section \ref{subsec:original_heuristic}) with negligible overhead (Section \ref{subsec:conceptual_discussion}).

\subsection{Experiment -- ADDRESS and MAPF-LNS}

\paragraph{Setting} We compare ADDRESS with the original MAPF-LNS using all stationary destroy heuristics of \cite{li2021anytime}, as described in Section \ref{subsec:mapf_lns_background}, for different time budgets on the \texttt{Den520d}, \texttt{Warehouse}, and \texttt{City} map with $m = 700$ agents. To evaluate the dominance of our ADDRESS heuristic over all stationary heuristics, we introduce a MAPF-LNS variant including all commonly used destroy heuristics, as well as our own.

\begin{figure*}
	\centering
	\includegraphics[width=0.9\textwidth]{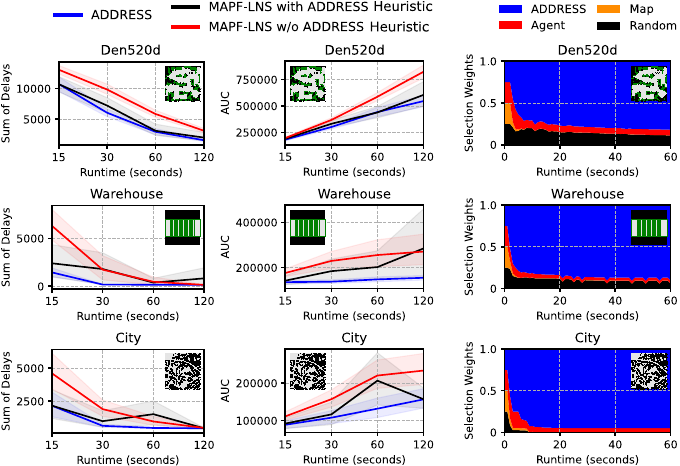}
     \caption{Sum of delays (\textbf{left}) and AUC (\textbf{middle}) for ADDRESS compared with the original MAPF-LNS (with and without our ADDRESS heuristic) for different time budgets (\underline{starting from 15 seconds}) with $m = 700$ agents in all maps. Shaded areas show the 95\% confidence interval. \textbf{Right}: Evolution of the selection weights of MAPF-LNS with our ADDRESS heuristic over time.}
     \label{fig:mapf_lns_runtime}
\end{figure*}

\paragraph{Results} The results are shown in Figure \ref{fig:mapf_lns_runtime}. ADDRESS outperforms both MAPF-LNS variants. The MAPF-LNS variant with our ADDRESS heuristic performs second best in \texttt{Den520d} and generally in the other maps with a maximum time budget of 30 seconds. Using our ADDRESS heuristics always leads to a lower average AUC when the time budget is lower than 120 seconds. The selection weights of MAPF-LNS indicate that our ADDRESS heuristic is the dominant destroy heuristic, as it is quickly preferred over all other heuristics. 

\begin{figure*}
	\centering
	\includegraphics[width=0.85\textwidth]{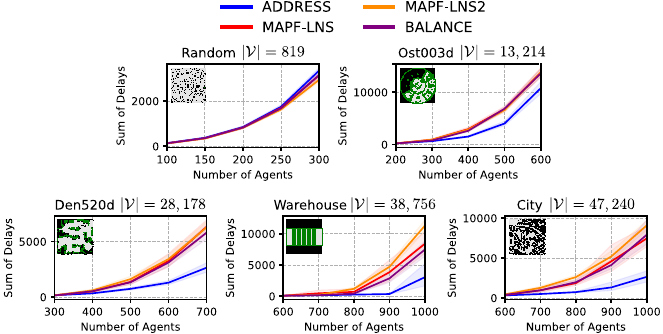}
     \caption{Sum of delays for ADDRESS compared with the original MAPF-LNS (without our ADDRESS heuristic), MAPF-LNS2, and BALANCE for different numbers of agents $m$ and a time budget of 60 seconds. Shaded areas show the 95\% confidence interval. The legend at the top applies across all plots. A comparison with LaCAM* is shown in Table \ref{tab:lacam_performance}. $|\mathcal{V}|$ denotes the corresponding map size, i.e., the number of occupiable locations.}
     \label{fig:address_results_star}
\end{figure*}

\paragraph{Discussion} The results confirm that our ADDRESS heuristic is more effective than the other heuristics in large-scale scenarios with $m = 700$ agents \cite{li2021anytime}, as it is consistently preferred by the original MAPF-LNS within less than 10 seconds of runtime. MAPF-LNS, with our ADDRESS heuristic, generally underperforms ADDRESS since it additionally explores the less effective destroy heuristics, whereas ADDRESS directly optimizes the seed agent selection for LNS neighborhood generation.

\subsection{Experiment -- State-of-the-Art Comparison}\label{subsec:exp3}

\paragraph{Setting} Finally, we compare ADDRESS with the original MAPF-LNS, MAPF-LNS2 (which finds feasible solutions by minimizing collisions), BALANCE, and LaCAM*. We run all algorithms on the \texttt{Random}, \texttt{Ost003d}, \texttt{Den520d}, \texttt{Warehouse}, and \texttt{City} maps with different numbers of agents $m$ and a time budget of 60 seconds.

\paragraph{Results} The results with ADDRESS, MAPF-LNS, MAPF-LNS2, and BALANCE are shown in Figure \ref{fig:address_results_star}. ADDRESS significantly outperforms all other approaches except in \texttt{Random}. BALANCE slightly outperforms MAPF-LNS and MAPF-LNS2 in \texttt{Den520d} and \texttt{Warehouse} with $m \geq 600$. Due to the large performance gap, we report the sum of delays of LaCAM* and ADDRESS separately in Table \ref{tab:lacam_performance} for the maximum number of agents per map tried in this experiment. ADDRESS (and all other baselines) clearly outperforms LaCAM*.

\begin{table}
\centering\small
\caption{Average sum of delays of ADDRESS and LaCAM* with 95\% confidence intervals with a time budget of 60 seconds and the maximum number of agents per map evaluated in Figure \ref{fig:address_results_star}. The best performance is highlighted in boldface.}
\begin{tabular}{|l||c|c|}\hline
          & ADDRESS & LaCAM* \\\hline\hline
\texttt{Random}    &  $\mathbf{3,343.52 \pm 120}$       &   $10,146.4 \pm 1399$     \\\hline
\texttt{Ost003d}   &  $\mathbf{10,788.4 \pm 1219}$       &   $38,632 \pm 3925$     \\\hline
\texttt{Den520d}   &  $\mathbf{2,646.4 \pm 433}$       &   $33,604.4 \pm 3823$     \\\hline
\texttt{Warehouse} &  $\mathbf{3,047.8 \pm 2165}$       &   $70,107.8 \pm 911$     \\\hline
\texttt{City}      &  $\mathbf{2,645.1 \pm 772}$   &   $48,760.6 \pm 614$    \\\hline
\end{tabular}\label{tab:lacam_performance}
\end{table}\normalsize

\paragraph{Discussion} The experiment demonstrates the ability of ADDRESS to outperform the state-of-the-art in large-scale scenarios with up to a thousand agents like in the \texttt{Warehouse} or \texttt{City} map. The high-level simplification of MAPF-LNS allows ADDRESS to focus its runtime on optimizing seed agents for neighborhood generation without \textbf{(1)} exploring less effective destroy heuristics or \textbf{(2)} iterating through the whole agent set $\mathcal{A}$, unlike the original agent-based destroy heuristic, used in MAPF-LNS and BALANCE. However, ADDRESS does not outperform the baselines in smaller scenarios, e.g., in the \texttt{Random} map. In this case, the overhead caused by agent sorting and Thompson Sampling outweighs the benefits of ADDRESS. In contrast, MAPF-LNS and BALANCE resort to the map-based heuristic, which is the dominant heuristic in the \texttt{Random} map  \cite{li2021anytime}.

\section{Conclusion}
We presented ADDRESS as a single-destroy-heuristic variant of MAPF-LNS. ADDRESS applies restricted Thompson Sampling to the top-$K$ set of the most delayed agents to select a seed agent for adaptive LNS neighborhood generation. Therefore, ADDRESS avoids time-consuming exploration of several stationary destroy heuristics.

Our experiments show that ADDRESS significantly outperforms state-of-the-art anytime MAPF algorithms like the original MAPF-LNS, MAPF-LNS2, BALANCE, and LaCAM* in large-scale scenarios with up to a thousand agents. The effectiveness of our destroy heuristic is confirmed by its lower costs and AUC compared with the original agent-based destroy heuristic in MAPF and the strong preference by the original MAPF-LNS over all other commonly used destroy heuristics. The combination of Thompson Sampling and the top-$K$ ranking of the most delayed agents enables efficient learning and a stronger focus on promising seed agent candidates through fast adaptation and filtering of agents whose paths were significantly shortened over time. ADDRESS with $\epsilon$-Greedy can also outperform state-of-the-art anytime MAPF with slightly weaker performance than Thompson Sampling, indicating that other MAB algorithms could be used, which we want to investigate in the future.

More future work includes the abstraction of agents and the application of our ADDRESS heuristic to other problem classes, such as TSP, MILP, or VRP, where variables can be sorted by their cost contribution to generate neighborhoods.

\section*{Acknowledgements}
The research at the University of Southern California was supported by the National Science Foundation (NSF) under grant numbers 1817189, 1837779, 1935712, 2121028, 2112533, and 2321786 as well as a gift from
Amazon Robotics. The views and conclusions contained in this document are those of the authors and should not be interpreted as representing the official policies, either expressed or implied, of the sponsoring organizations, agencies, or the U.S.  government.

\bibliography{references}

\end{document}